%% file: main.tex
\documentclass{ws-ijairr}
\usepackage[super,sort&compress,comma]{natbib}
\usepackage{url}
\usepackage{subcaption}
\usepackage{array}
\usepackage{xcolor,pifont}
\newcommand*\colourcheck[1]{%
  \expandafter\newcommand\csname #1check\endcsname{\textcolor{#1}{\ding{52}}}%
}
\newcommand*\colorcross[1]{%
  \expandafter\newcommand\csname #1cross\endcsname{\textcolor{#1}{\ding{56}}}%
}
\colourcheck{green}
\colorcross{red}
\begin{document}

% \markboth{F. Author \& S. Author (authors' names)}{Instructions for typing manuscripts (paper's title)}

\markboth{M. S. Sakib \& Y. Sun}{Cooking Recipe to Robot Task Tree}

\catchline{}{}{}{}{}
%
%%%%%%%%%%%%%%%%%%%%%%%%%%%%%%%%%%%%%%%%%%%%%%%%%%%%%%%%%%%%%%%%%%%%

% \title{Instructions for Typesetting Manuscripts\\
% using \LaTeX\footnote{For the title, try not to
% use more than 3 lines. Typeset the title in 10~pt
% Times Roman, boldface, and \textit{title case}: first letter of important words to be capitalized.}}
%\title{Enhancing Task Planning Accuracy in Large Language Models: Generating PDDL Plans from Natural Language Instructions}
\title{Consolidating Trees of Robotic Plans Generated Using Large Language Models to Improve Reliability}

% \author{Md Sadman Sakib\footnote{Typeset names in 8~pt Times Roman, upper and lower case.
% Use the footnote to indicate the present or permanent address of
% the author.}}
\author{Md Sadman Sakib}

\address{Department of Computer Science, University of South Florida, \\
Tampa, FL, USA\\
mdsadman@usf.edu}

\author{Yu Sun}

\address{Department of Computer Science, University of South Florida, \\
Tampa, FL, USA\\
yusun@usf.edu}

\maketitle

% \begin{history}
% \received{Day Month Year}
% \revised{Day Month Year}
% \accepted{Day Month Year}
% \end{history}

\begin{abstract}

The inherent probabilistic nature of Large Language Models (LLMs) introduces an element of unpredictability, raising concerns about potential discrepancies in their output. 
This paper introduces an innovative approach aims to generate correct and optimal robotic task plans for diverse real-world demands and scenarios. LLMs have been used to generate task plans, but they are unreliable and may contain wrong, questionable, or high-cost steps. The proposed approach uses LLM to generate a number of task plans as trees and amalgamates them into a graph by removing questionable paths.
%, reinforcing valid paths, and choosing low-cost alternatives over high-cost paths. 
Then an optimal task tree can be retrieved to circumvent questionable and high-cost nodes, thereby improving planning accuracy and execution efficiency. The approach is further improved by incorporating a large knowledge network. 
%This paper introduces an innovative task planning pipeline dedicated to enhancing LLM performance in generating Planning Domain Definition Language (PDDL) plans for robotic tasks from natural language instructions. Our approach involves crafting multiple high-level task plans, depicted as task trees, employing prompt engineering methods using GPT-4 to capture intricate sequential and parallel dependencies among subtasks. These task trees are amalgamated into a network to address uncertainties and common unreliable features in LLM outputs. Moreover, the pipeline utilizes a task tree retrieval mechanism to circumvent questionable and high-cost nodes, thereby improving planning accuracy and execution efficiency. 
Leveraging GPT-4 further, the high-level task plan is converted into a low-level Planning Domain Definition Language (PDDL) plan executable by a robot. Evaluation results highlight the superior accuracy and efficiency of our approach compared to previous methodologies in the field of task planning.

\end{abstract}


\input{1-introduction}

\input{2-data-structure}

\input{3-task-tree-generation}
\input{4-pddl-generation}

\input{5-experiments}    
\input{6-discussion}    
\input{7-conclusion}

\bibliographystyle{ws-ijairr}
\bibliography{ref}

\end{document}

%% file: 1-introduction.tex
\section{Introduction}

Navigating real-world tasks using robots often proves to be a challenge that traditional task planning systems struggle to overcome. These methodologies heavily rely on having complete domain knowledge \cite{Xu_Wang_Niu_Wu_Che_2020, baier, Nau}, which is rarely feasible in dynamic real-world settings. 
A typical instance illustrating the limitations of such knowledge-based approaches can be observed when a traditional system is asked to generate a plan to prepare a dish whose recipe does not exist in its knowledge base. It will inevitably fail to devise a plan as the necessary information to carry out that task is not available.
Such circumstances expose the inherent shortcomings of traditional search-based planning methods in dealing with unforeseen scenarios.

The emergence of Large Language Models (LLMs) has broadened the scope of problem-solving \cite{YANG2023100007, Guo2022FromIT, zhang2023bootstrap}, presenting potential solutions for various scenarios and inquiries. Nevertheless, their inherent probabilistic nature occasionally generates outputs lacking consistent accuracy or optimality. This divergence between their potential and actual performance emphasizes the crucial demand for innovative methodologies to enhance and strengthen the effectiveness of LLMs in task planning.

\begin{figure}[ht]
	\centering
	\includegraphics[width=\columnwidth]{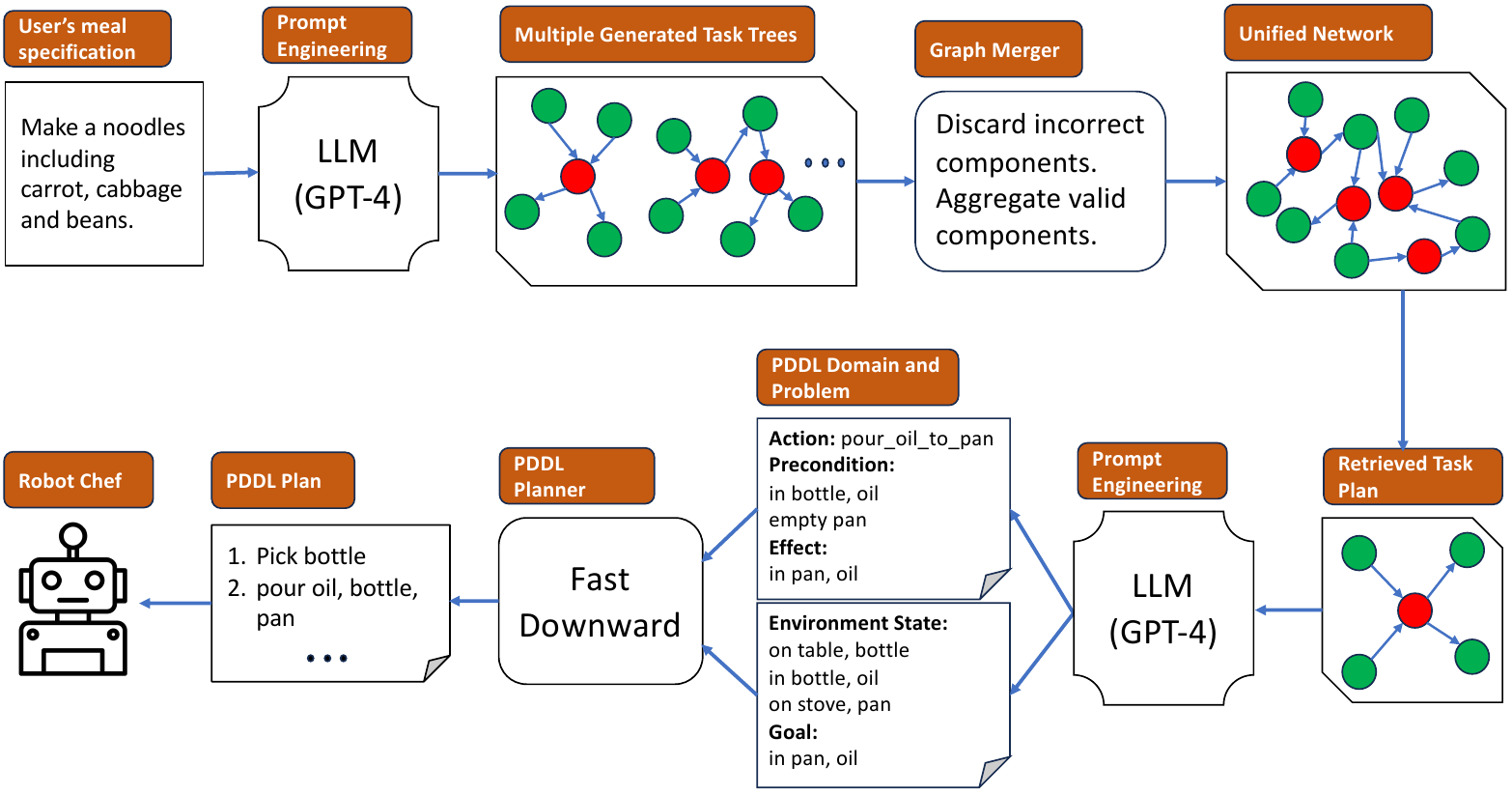}
	\caption{Overview of our task planning approach (using a cooking task for example). The process begins with a meal specification query, resulting in the creation of an optimal task tree. This tree is then converted into a PDDL plan, facilitating robot task execution.}
	\label{fig:pipeline}
\end{figure}

This study introduces a customized framework aimed at boosting the capabilities of LLMs in the domain of robotic task planning. The primary objective is to transform natural language directives into Planning Domain Definition Language (PDDL)\cite{mcdermott1998pddl, DBLP:conf/aaai/HollerBBBFPA20, fox2003pddl2} plans (Figure~\ref{fig:pipeline}). Instead of the traditional technique that produces only one plan, our approach results in multiple high-level task plans shaped into task trees, leveraging GPT-4. Parallel or sequential complexities among subtasks are captured by these task trees. Afterward, we combine these task trees into a unified network, eliminating the unreliable components from it. This network may offer multiple viable solutions, some potentially entailing costly nodes challenging for the robot's execution. To address this, we implement a task tree retrieval strategy that strategically bypasses these resource-intensive nodes. This significantly elevates planning accuracy and overall task execution efficiency. 

The high-level plan may not be suitable for direct robotic execution. It requires grounding each high-level subtask with specific low-level skills. Leveraging the capabilities of GPT-4, the high-level task plan seamlessly translates into a low-level PDDL plan. This detailed plan is optimized for implementation by robotic systems.

The effectiveness of our method is assessed through a comparative analysis against prior methodologies. Our experimentation involves creating task plans for a cooking robot. The outcomes exhibit the superiority of the suggested method, highlighting improved task planning accuracy and enhanced cost-efficiency. Though our approach is not confined to a specific field, in this paper, we predominantly utilize cooking examples to facilitate better understanding and visualization.

In summary, this paper introduces an innovative method that significantly enhances the reliability of Language Model-based Learning (LLMs) for generating robotic task plans. This is achieved in three steps. First, we trained an LLM-based model that generates trees of robotic plans from a given demand. Then, we developed an amalgamation algorithm that effectively eliminates questionable paths and replaces high-cost alternatives with more efficient, low-cost options. Lastly, we developed a search algorithm that ensures the selection of the most optimal tree for the given task. The performance of the proposed approach can be further improved by allowing users to make corrections using the easily comprehensible visualization of the task tree before converting it into machine-friendly PDDL code.

%In brief, our key contributions are: - innovative way of improving reliability of LLMs for generating robotic task plans through consolidating trees of robotic plans;- a set of algorithms that enable it: generate a good set of trees; amalgamation algorithm that removes questionable paths, and replacing high-cost paths with low-cost alternatives; search algorithm to get an optimal tree; interface to allow user to modify before convert it to unreadable PDDL; convert trees to PDDL domain file/problem file. 

% (i) Introducing a pipeline that transforms human instructions into a PDDL plan; (ii) Proposing an approach to enhance the efficiency of LLMs in robot task planning; (iii) Providing a method to visualize the plan at an intermediate stage to gather human feedback if necessary; (iv) Displaying the superiority of our model through comparison with a previous approach. 

% In the following sections, we will delve into the details of the task tree generation pipeline, including the representation of the task tree structure, the integration of ChatGPT, GPT-3 fine-tuning, and the utilization of FOON for cost optimization. We will also present the experimental evaluation and discuss the implications of our findings. Through this research, we aim to contribute to the advancement of robotic cooking and facilitate the integration of robots into our culinary routines.

\subsection{Related Works}

In the domain of task planning, several strategies have been proposed to tackle the challenges associated with generating effective action sequences for executing user instructions. One prominent approach involves the use of knowledge graphs to address this challenge \cite{paulius2019survey}. Notably, the KNOWROB framework \cite{Beetz2018KnowR2, pancake} has made significant contributions in this area by leveraging a knowledge base constructed from data collected in sensor-equipped environments. Daruna et al.\cite{Daruna2021TowardsRO} introduced a task generalization scheme that relaxes the requirement of having multiple task demonstrations to perform tasks in unknown environments. This scheme integrates the task plan with a knowledge graph derived from observations in a virtual simulator. The impact of knowledge graphs on a robot's decision-making process was further investigated \cite{Daruna2022ExplainableKG}. However, these approaches heavily rely on the limited information contained in their respective knowledge bases. In contrast, our approach harnesses the power of LLM to alleviate the burden of creating a knowledge base, offering a more comprehensive and flexible solution.

In recent research, there has been a growing focus on task planning with LLMs, capitalizing on the remarkable language comprehension and generation capabilities these models possess. Several studies have explored the use of LLMs to create step-by-step plans for long-horizon tasks. For instance, Erra \cite{erra} proposed an approach employing LLMs for generating plans in complex manipulation tasks. Moreover, numerous research endeavors, such as those by Huang et al. \cite{Huang2022LanguageMA}, Shah et al. \cite{Shah2022LMNavRN}, and Zeng et al. \cite{Zeng2022SocraticMC}, have investigated plan generation using LLMs across various domains.
% While these works have made valuable contributions to plan generation for robots, they often
However, these works often do not explicitly consider the robot's capability to perform specific actions. One limitation of relying solely on LLMs is the lack of interaction with the environment. To address this limitation, SayCan \cite{saycan} introduced a framework that combines the high-level knowledge of LLMs with low-level skills, enabling the execution of plans in the real world. By grounding LLM-generated plans with the robot's capabilities and environmental constraints, SayCan bridges the gap between language-based planning and physical execution.
In addition, recent research efforts such as Text2Motion \cite{lin2023text2motion}, ProgPrompt \cite{singh:progprompt} have integrated LLMs with learned skill policies. 
% \cite{singh:progprompt} employs prompt engineering instead of fine-tuning the model, which works well for simple tasks but may face challenges when applied to complex tasks like cooking.
They exhibit trust in the LLM-generated plan and proceed to execute it whereas we focus on enhancing LLM's accuracy to generate an optimal task plan.
A recent study, LLM+P \cite{LLMP}, converts user commands into a PDDL problem definition and subsequently utilizes a classical planner to obtain a plan. However, a significant concern with their approach is the assumption of the existence of a pre-defined PDDL domain file. This domain file is not always available, leading to planning failures due to insufficient information in the knowledge base.

New research is exploring the integration of Large Language Models (LLMs) with Knowledge Graphs (KGs), combining the strengths of both tools \cite{meyer2023llm}. LLMs offer capabilities for language processing and generalisation, while KGs offer structured, precise, and sector-specific information. Pan et al. examined various models for unifying LLMs and KGs \cite{pan2023unifying}, while Zhu et al. examined how well LLMs perform when constructing KGs \cite{zhu2023llms}.
One common problem with LLMs is the production of incorrect or non-existent answers, known as hallucinations \cite{Ji2022SurveyOH, Bang2023AMM, Wang2023SurveyOF}. However, recent studies have shown that the use of external knowledge resources can reduce the occurrence of these hallucinations and enhance the performance of LLMs \cite{agrawal2023can, zheng2023does, ge2023openagi}.

Table \ref{table:related_work_summary} compiles a comparative analysis of specific key elements present in various methodologies. These elements encompass (i) the integration of LLM as a knowledge repository, (ii) adaptability to function in information-deficient environments, (iii) the ability to display reasoning steps or intermediate stages within the pipeline output, (iv) the existence of mechanisms to improve planning precision through automated processes or user feedback, and (v) a specific focus on optimizing task plans by reducing execution costs. Despite the diversity in their approaches, all these strategies share the common trait of accepting natural language input instructions and generating a  task plan.

\begin{table}[ht]

\centering
% \begin{tabular} {|p{0.16\linewidth}|p{0.16\linewidth}|p{0.16\linewidth} | p{0.16\linewidth} | p{0.16\linewidth}| p{0.16\linewidth}|p{0.16\linewidth}|}
% \hline
\begin{tabular} {p{0.15\linewidth} >{\centering\arraybackslash} p{0.15\linewidth} >{\centering\arraybackslash}p{0.14\linewidth} >{\centering\arraybackslash} p{0.11\linewidth}  >{\centering\arraybackslash} p{0.11\linewidth} >{\centering\arraybackslash} p{0.13\linewidth}}
\hline

 & \textbf{LLM as Knowledge Base?} & \textbf{Operate in Unseen Environment?} & \textbf{Plan Visualization?} & \textbf{Plan Correction?} & \textbf{Optimize task plan?} \\ \hline
FOON\cite{paulius2016functional} & \redcross & \redcross  & \redcross & --- & \redcross \\
W-FOON\cite{paulius2021weighted} & \redcross & \redcross  & \redcross & --- & \greencheck \\
FOON+S\cite{Sakib2022ApproximateTT} & \redcross & \greencheck  & \greencheck & \redcross & \redcross \\
Dual-arm\cite{dualarm} & \redcross & \redcross  & \redcross & \greencheck & \greencheck \\
SayCan\cite{saycan} & \greencheck & \greencheck  & \redcross & \redcross & \greencheck \\
RTPO\cite{rtpo} & \greencheck & \greencheck  & \redcross & \redcross & \redcross \\
ProgPrompt\cite{singh:progprompt} & \greencheck & \greencheck  & \greencheck & \greencheck & \redcross \\
LLM+P\cite{LLMP} & \redcross & \greencheck  & \redcross & --- & \redcross \\ \hline 
Ours & \greencheck & \greencheck  & \greencheck & \greencheck & \greencheck \\ \hline

\end{tabular}
\caption{Comparative analysis of key features: our approach versus related works.}
\label{table:related_work_summary}
\end{table}

%% file: 2-data-structure.tex
\section{Plan Representation}

Our approach constitutes use of a tree-structured data representation, designated as a ``task tree", for showcasing task plans. The specific elements that compose this framework are expanded upon in the following subsections.

\subsection{Components of the Task Tree}
A task tree comprises two distinct nodes: object nodes and motion nodes.
For the purpose of illustration, let's look at a cooking scenario. Object nodes denote an array of relevant items, such as ingredients (e.g., vegetables, condiments), kitchen utensils (e.g., knives, spoons), containers (e.g., bowls, cutting boards), and kitchen appliances (e.g., stoves, ovens). Every object node encompasses a unique subset of attributes, which include the object's label, its current physical state, its location, and—particularly for composite objects or containers—the composition of ingredients. For instance, a scrambled egg—a composite object—consists of a variety of ingredients.
Conversely, motion nodes encapsulate actions that engage with objects, spanning from high-level tasks like pouring or mixing to low-level tasks such as grasping or pushing.

A motion node is connected to an object node through a directed edge. Inbound edges to a motion node connect the necessary object nodes for performing the action, while outbound edges from the motion node link to the object nodes resulting from the action.

\subsection{Functional Unit}

\begin{figure}[!ht]
     \begin{subfigure}[t]{\textwidth}
         \includegraphics[width=\textwidth]{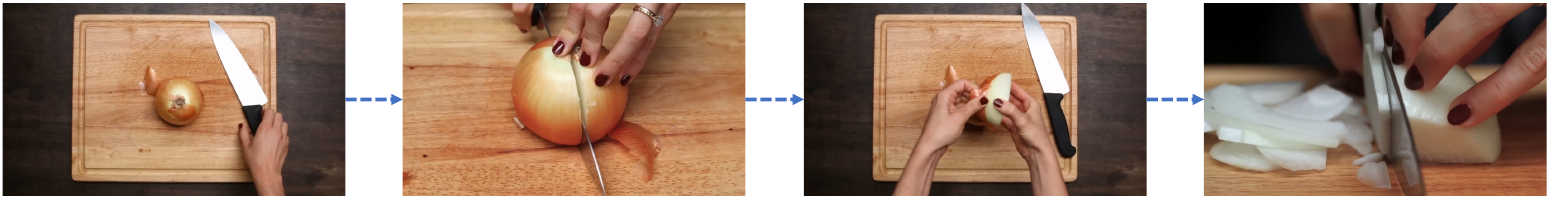}
         \caption{}
         \label{fig:video}
     \end{subfigure}
     \begin{subfigure}[t]{\textwidth}
         \includegraphics[width=\textwidth]{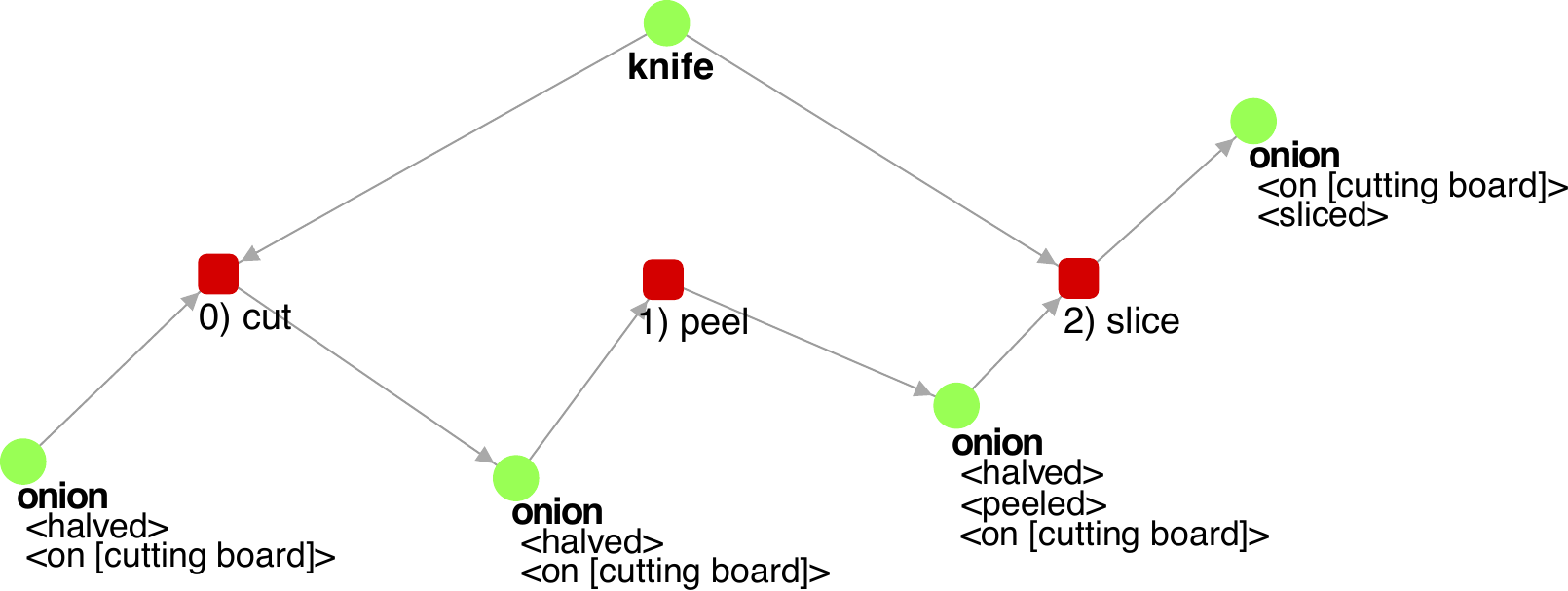}
         \caption{}
         \label{fig:fu}
     \end{subfigure}
     \caption{(a) Activities from a cooking video and (b) its corresponding functional units.}
    \label{fig:fu-creation}
\end{figure}

A functional unit signifies a singular action within the plan, akin to a planning operator in PDDL. Input object nodes within it represent preconditions, while output object nodes denote effects. It comprises at least one input object, at least one output object, and precisely one motion. 
In Figure~\ref{fig:fu-creation}, we show an example of how the steps in a cooking video of slicing an onion can be represented using functional units. 

% \subsubsection{Subgraph}

% The fundamental building blocks of FOON are functional units. These units allow us to vividly describe actions observed in demonstrations. A sequence of functional units describing a single activity, recipe, or demonstration is referred to as a FOON "subgraph." Subgraphs are typically constructed by annotators who observe instructional videos and document the objects and actions in the form of functional units.

% As of the present, our FOON dataset comprises 140 subgraphs, corresponding to recipe and demonstration videos from sources like YouTube, Activity-Net, and EPIC-KITCHENS. These annotations, along with code for processing the data, are available online.

% \subsubsection{Universal FOON}

% A comprehensive knowledge graph, termed a "universal FOON," is constructed by amalgamating two or more subgraphs. This amalgamation process involves identifying duplicate functional units, units that share the exact combination of input and output object nodes and motion nodes. A universal FOON provides a consolidated representation of various ways to prepare ingredients or meals, as well as common functional units across diverse recipes.

\subsection{Task Tree}

A task tree is just a representation of all the step to complete a task. It represents a sequence of functional units, aimed at achieving a defined goal or target object (e.g., a finished meal in the context of cooking) from an initial environmental state. This task tree guides the robot in executing a series of actions to attain the desired outcome.

%% file: 3-task-tree-generation.tex
\section{High-level Task Tree Generation}
During this stage, a high-level task tree for robotic operations is produced based on the user command. This process involves utilizing an LLM to generate multiple potential task plans organized as task trees. These plans are amalgamated to form a comprehensive knowledge network. Next, a graph search method is applied to extract a more precise and cost-effective task tree from this unified network. Integration of an external knowledge source into this pipeline is possible to enhance the task tree's quality further.

\subsection{Creation of potential task trees}
\label{multiple_tree}
\subsubsection{Procedure for Generating Instruction Steps}
\label{instructions}
Users have the freedom to specify diverse criteria in their commands, such as dietary requirements (e.g., gluten-free, vegetable-based, or non-dairy preferences). An example of a typical user command could be: ``Prepare a soup with carrots, cabbage, and beans, mildly spiced." From this input, we formulate a prompt and extract the recipe using GPT-4. Alternative LLMs such as GPT-3\cite{Brown2020LanguageMA} or ChatGPT\cite{chatgpt} can also be utilized in this step without adversely impacting the overall performance of our approach. To leverage the output for the next phase of our pipeline, we create the prompt to guide the model in producing itemized instructions within its response, explicitly instructing it to articulate the instructions in steps. However, the resultant output is in natural language, potentially posing challenges for direct execution by a robot.    

\subsubsection{Transformation of Instruction Steps into a Task Tree}
When a robot performs an action, several factors need to be considered, such as preconditions, effects, and the objects involved. Additionally, understanding the state of these objects is crucial for determining the appropriate grasp or manipulation technique. However, extracting all this information from a recipe written in natural language poses significant challenges. To address this complexity, we translate the instructions into structured functional units that encapsulate all the necessary details.
These functional units are then organized into a task tree, providing a systematic guide for the robot to effectively execute the task.

\begin{figure}[ht]
	\centering
	\includegraphics[width=\columnwidth]{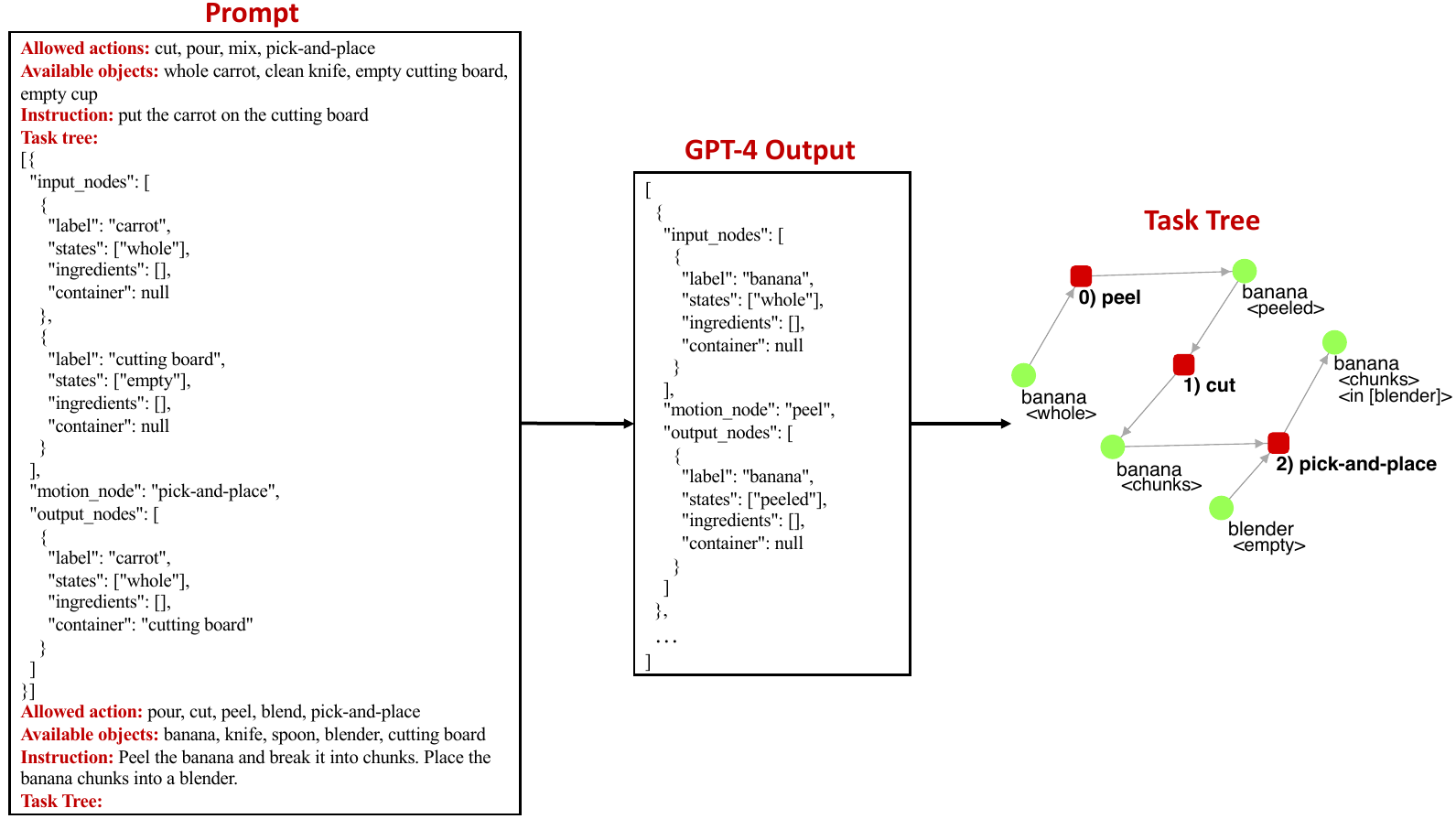}
	\caption{An example of task tree generation from user command using prompt engineering with GPT-4.}
	\label{fig:progmpt-eng}
\end{figure}

To achieve this, we employ prompt engineering using GPT-4. Within the prompt, we include examples of generated task trees for actions and items a robot can utilize, along with the instructions it receives. We then ask GPT-4 to generate a task tree for the instructions outlined in the previous section. The resultant task tree is presented in a JSON format, comprising several functional units.
The simplified version of the prompt utilized in our approach is depicted in Figure~\ref{fig:progmpt-eng}.

\subsection{Creation of a unified network}
\label{unified}
To mitigate potential errors in the task plans generated during the previous phase, our approach involves the generation of multiple task trees. The goal is to identify an error-free and efficient task tree for robot execution from the combined graph. Previous studies like FOON\cite{paulius2016functional} and EvoChef\cite{Jabeen2019EvoChefSM} have demonstrated that merging recipes can lead to the discovery of innovative cooking methods by allowing recipes to share information and learn diverse subtask approaches. Drawing inspiration from this concept, we create multiple recipes and their corresponding task trees using the method outlined in section \ref{multiple_tree}. During our experimentation, we observed that generating five distinct trees for a particular task and combining them into a network yielded encouraging results. 

Throughout the merging process, our primary objective is to eliminate incorrect functional units and prevent duplicate entries. Incorrect functional units may arise from either syntax errors or erroneous object-action relationships. Syntax verification entails confirming the presence of essential components in each functional unit, such as input and output objects, as well as a motion node. Furthermore, it involves validating whether each object has an assigned state. Assessing the object-action relationship correctness poses the challenge of determining if the action's state transition is accurate. To address this challenge, we have compiled a comprehensive list of valid state transitions from FOON. This list serves as the basis for assessing the accuracy of a transition. For example, if a transition such as ``chopped $\rightarrow$ whole" is absent in FOON, it is identified as incorrect. Functional units that successfully meet these verification criteria are subsequently integrated into the network.

\subsection{Task tree retrieval from the unified network}
Using the desired dish as the targeted node, we deploy a search technique similar to Weighted FOON \cite{paulius2021weighted} to explore all available paths leading to this objective. We consider the node representing the desired output as the goal node and the initial states of the objects as the leaf nodes within the network. Employing Depth-First Search (DFS), we extract paths from these leaves to reach the goal. This exploration often results in multiple task plans showcasing novelty, varying in steps or object manipulations. For instance, in the scenario of preparing a strawberry smoothie, one plan might involve adding the whole strawberry to the blender, while another could suggest slicing the strawberry before blending.

After filtering out erroneous functional units, our task tree retrieval procedure avoids their selection, giving precedence to accurate functional units in constructing the task plan. For example, if the ``pouring milk to cup" functional unit is deemed incorrect in the first generated tree but correct in the other four task trees, the search procedure will favor the accurate functional unit from those remaining four trees.

\subsection{Retrieving Cost-Effective Trees}
From the assortment of generated plans, it is crucial to select one that the robot can feasibly carry out. The robot's configuration largely dictates this feasibility. A single-handed robot, for example, may find the act of pouring simpler than chopping. As such, the success rate of a task tree's execution can differ among diverse robots. In line with the methodology by Weighted FOON \cite{paulius2021weighted}, we allocate a cost value ranging between 0 and 1 to each action. These respective values are influenced by three key factors: 1) the robot's physical abilities, 2) its historical experiences and competency in executing actions, and 3) the objects or tools the robot needs to interact with. For our experiment, we have assigned the cost based on our experience working with the UR5 robot. 
A higher cost value signifies an action that is more challenging to execute.

\begin{figure}[ht]
     \begin{subfigure}[t]{0.49\textwidth}
         \includegraphics[width=\textwidth]{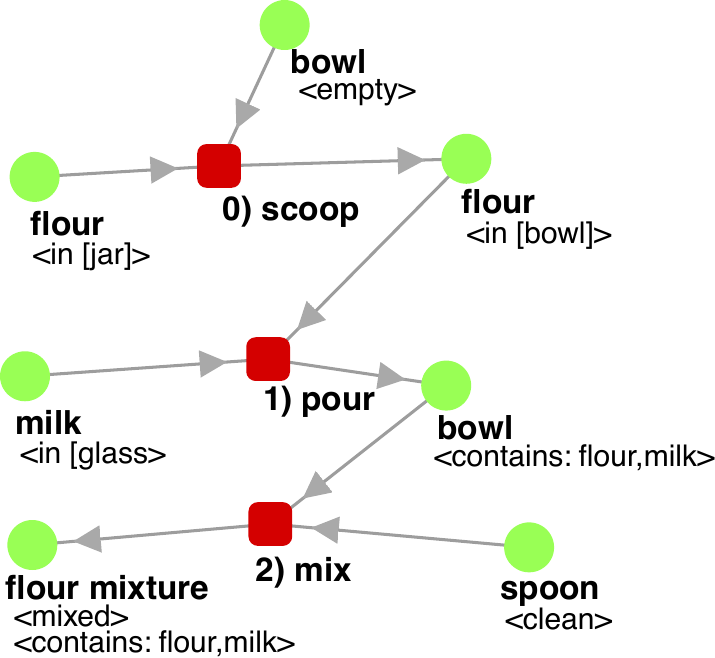}
         \caption{cost of execution = 0.7}
         \label{fig:scoop1}
     \end{subfigure}
     \begin{subfigure}[t]{0.49\textwidth}
         \includegraphics[width=\textwidth]{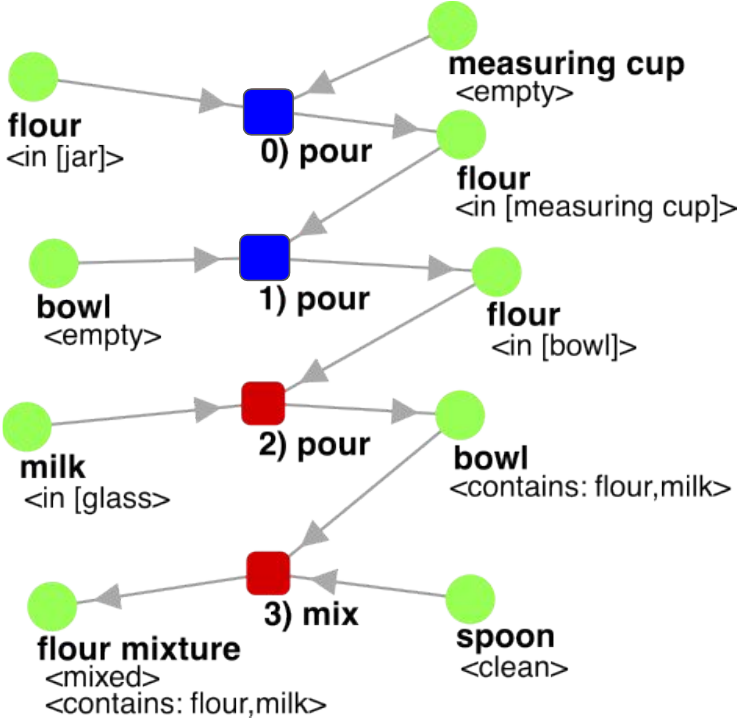}
         \caption{cost of execution = 0.5}
         \label{fig:scoop2}
     \end{subfigure}
     \caption{Illustration of cost optimization: A comparison between task trees obtained from (a) GPT-4 and (b) the unified network. The assigned costs for scooping, pouring, and mixing are 0.4, 0.1, and 0.2, respectively.}
    \label{fig:example}
\end{figure}

Figure \ref{fig:example} showcases an example of cost optimization taking advantage of the unified network, which prioritizes two pouring actions over scooping due to the significantly lower associated cost (0.1) as opposed to scooping (0.4). The rationale for assigning a reduced cost to pouring was influenced by the successful pouring accuracy demonstrated by Huang et al. \cite{huang2017learning,HUANG2021103692} using the UR5 robot.

\subsection{Integration of external knowledge resources}

\begin{figure}[!ht]
     \begin{subfigure}[t]{0.49\textwidth}
         \includegraphics[width=\textwidth]{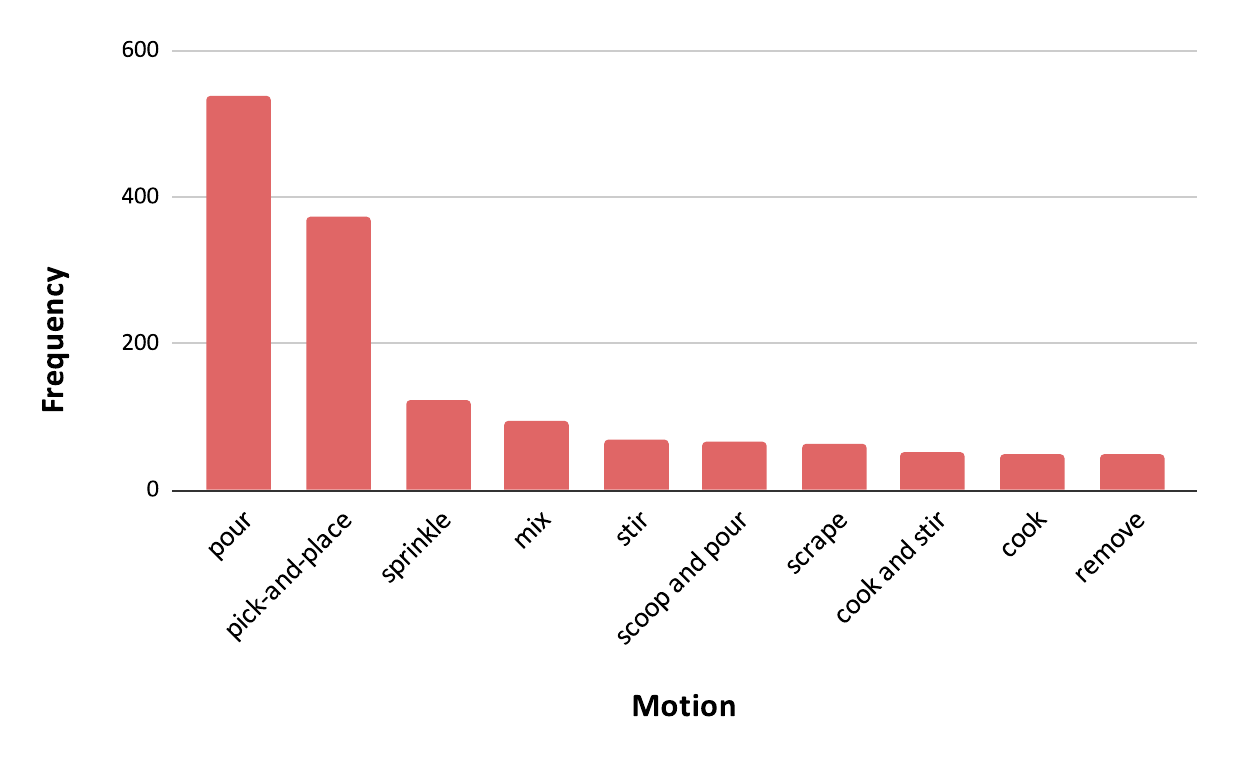}
         \caption{}
         \label{fig:motion}
     \end{subfigure}
     \begin{subfigure}[t]{0.49\textwidth}
         \includegraphics[width=\textwidth]{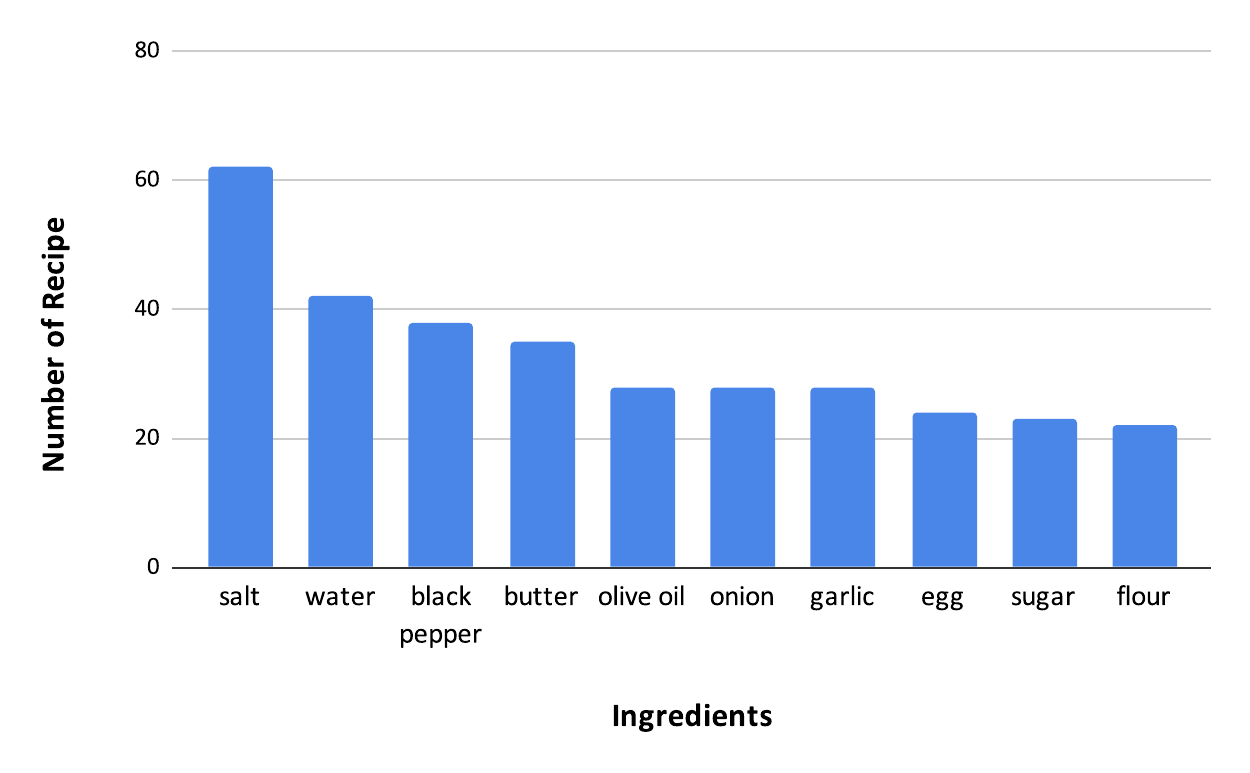}
         \caption{}
         \label{fig:ingredients}
     \end{subfigure}
     \begin{subfigure}[t]{0.49\textwidth}
         \includegraphics[width=\textwidth]{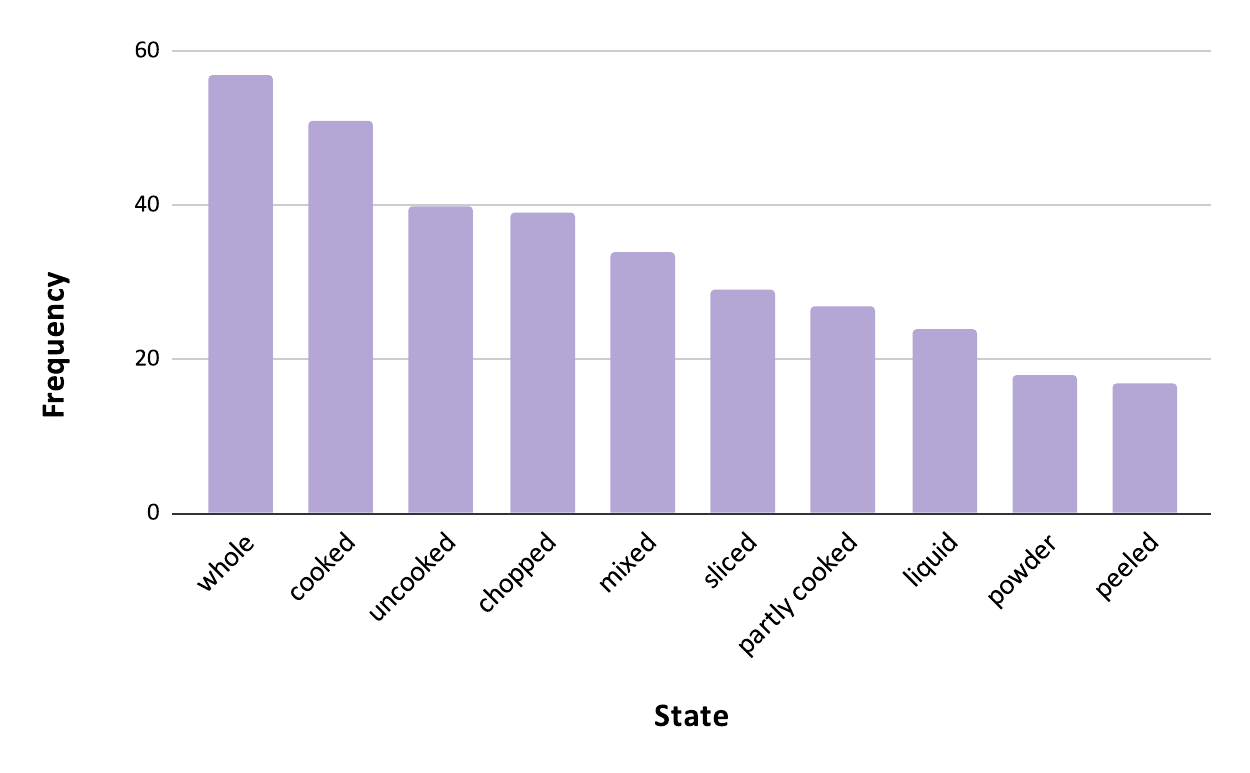}
         \caption{}
         \label{fig:state}
     \end{subfigure}
     \begin{subfigure}[t]{0.49\textwidth}
         \includegraphics[width=\textwidth]{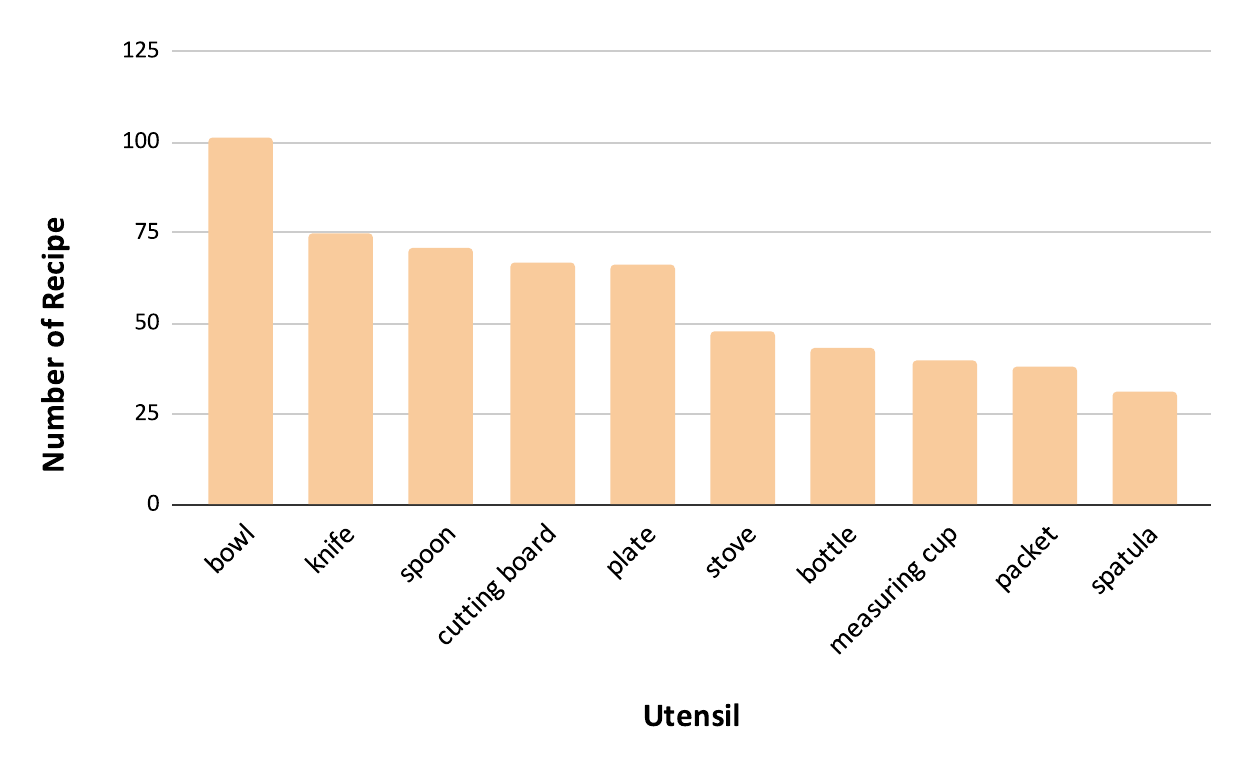}
         \caption{}
         \label{fig:utensil}
     \end{subfigure}
     \caption{Top 10 most frequently seen \textbf{(a)} motions, \textbf{(b)} ingredients, \textbf{(c)} states and \textbf{(d)} utensils in FOON.}
    \label{fig:stat}
\end{figure}

Our pipeline is designed to easily incorporate an external knowledge base, which can be expressed via functional units and task trees, thereby enhancing the quality of generated plans. Given the context of our experiments within the culinary domain, we utilize FOON\cite{paulius2016functional,paulius2018functional}. This resource comprises 150 cooking recipes, each represented as a task tree. 
Each recipe was individually annotated by a person who watched the cooking video, generating a corresponding task tree similar to the one illustrated in Figure \ref{fig:fu}. Subsequently, another individual evaluated and corrected the tree. This comprehensive review and correction process were undertaken to diminish any noise present in the dataset.
% This dataset, curated manually, was developed by annotators who watched cooking videos and created corresponding task trees in a manner akin to what is shown in Figure \ref{fig:fu}. 
All these task trees were consolidated into a single unified network to formulate FOON. Figure \ref{fig:stat} provides an overview of the most common motions \cite{paulius2020motion, huang2019dataset}, ingredients, utensils, and states of objects encountered in this network. 

In order to integrate FOON into our pipeline, each of its functional units is incorporated into the unified graph, such that it comprises plans from both GPT-4 and FOON. 
During this merging process, our primary focus lies in the consolidation of nodes. Both GPT-4 and FOON may label the same object or motion node differently. To harmonize these disparities, we follow a series of elementary rules, including converting all object names to their singular form. This addresses discrepancies such as GPT-4 often generating plurals like ``strawberries" and ``onions", whereas FOON uses ``strawberry" and ``onion" respectively. The enlargement of the network likely enhances our capability to derive more task trees. An additional advantage is the increased likelihood of pinpointing a more accurate and cost-efficient plan, given FOON's meticulous curation specifically tailored to robotic execution. 

Our experiments have shown that this larger network yields more cost-effective trees compared to a knowledge base that excludes FOON.
However, the expansion of the network also escalates the retrieval time for all task trees. Nevertheless, this time complexity can be mitigated through heuristic-based searches such as the approximate tree retrieval approach \cite{Sakib2022ApproximateTT} or Rapidly-Exploring Random Tree Star (RRT*) \cite{rrt}.

%% file: 4-pddl-generation.tex
\section{Task Tree to PDDL Conversion}
A task tree provides a high-level plan that lacks interaction with the environment.
However, executing actions often requires additional information, such as the geo-
metric position of objects or the initial quantity of ingredients in a container. For
instance, a task plan might involve adding ice to an empty glass, but the glass could
be positioned upside down on a table. Therefore, before pouring the ice, the glass
would need to be rotated back to its original position. This crucial step is missing
in the high-level planning. Hence, there is a need for hierarchical planning, where
the task tree can be converted into a low-level plan that can be executed in the real
world.

\begin{figure}[ht]
	\centering
	\includegraphics[width=\columnwidth]{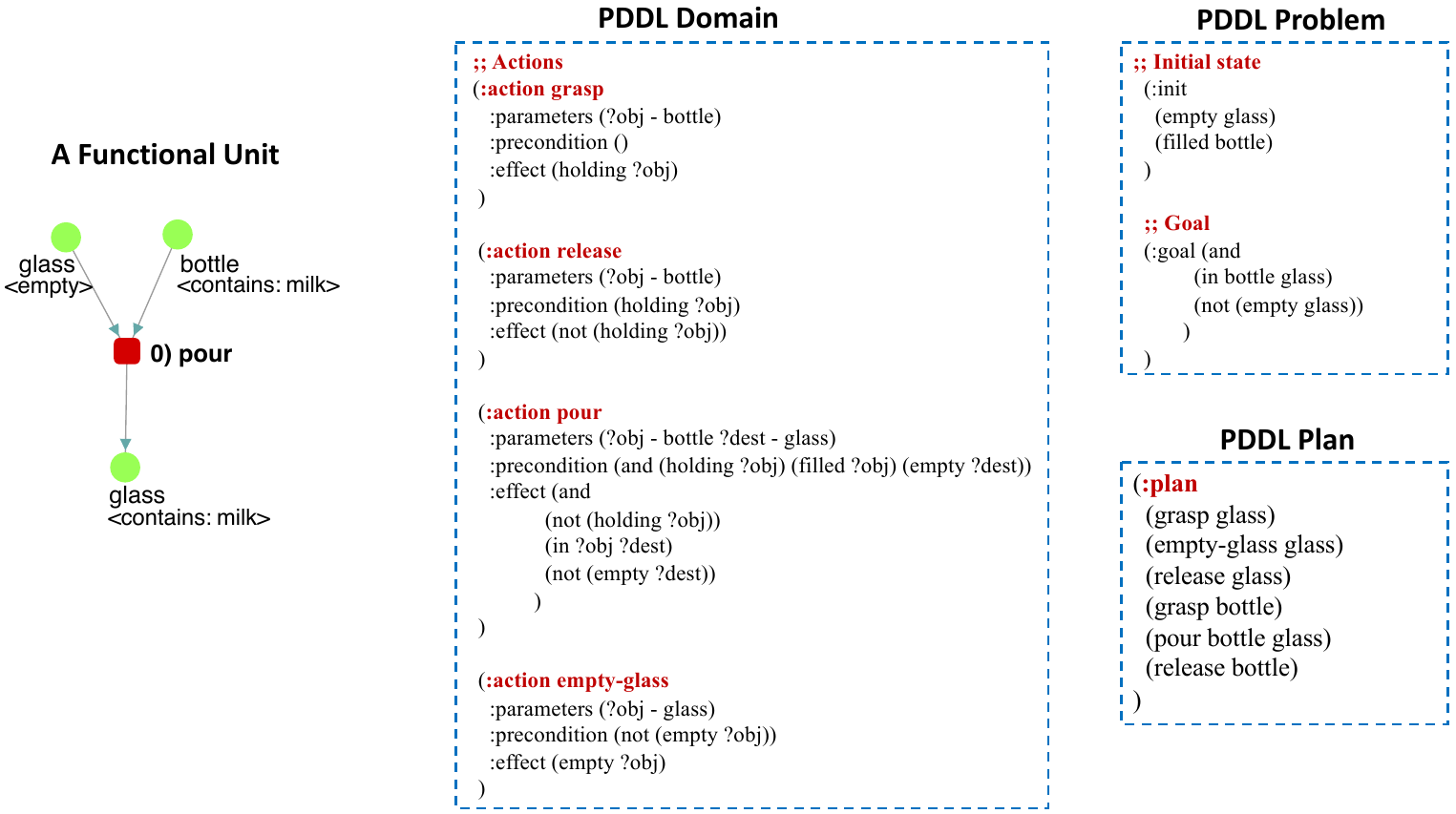}
	\caption{PDDL Plan Generation for Pouring Milk from Bottle to Glass. Initially, GPT-4 is employed to create both the domain and problem files, followed by the utilization of a PDDL planner to discover the plan.
}
	\label{fig:pddl}
\end{figure}

In the subsequent stages of our pipeline, we proceed to transform the task tree into a PDDL plan encompassing detailed low-level robot skills.
Although Paulius et al. \cite{Paulius2022LongHorizonPA} introduced a method to derive the PDDL plan from a task tree, this approach has limitations. It necessitates the manual definition of all robot-executable primitives, making it a highly demanding and error-prone task. Furthermore, manually formulating the problem in PDDL adds to the potential for errors. This conversion process lacks easy adaptability to different domains.
Conversely, LLMs are well-acquainted with PDDL. Hence, in this phase, we employ GPT-4 once more. Each functional unit is treated as an autonomous problem, allowing us to generate PDDL domain and problem files. While converting a functional unit into low-level actions, we include in the prompt which actions can be executed by the robot, ensuring that the transformed plan does not comprise operations that the robot is unfamiliar with. Recent advancements \cite{LLMP, xie2023translating} have demonstrated the ease of formulating PDDL problems using natural language through LLM.

However, recent evidence strongly suggests that relying solely on LLMs for planning is not ideal \cite{Collins2022StructuredFA, Valmeekam2022LargeLM, Mahowald2023DissociatingLA}.
Multiple classic PDDL planners \cite{Hoffmann, baier, Helmert2006TheFD} are available, ensuring highly accurate plan derivation from the domain and problem files. For our purposes, we employ the Fast-Downard planner to generate the plan. Figure \ref{fig:pddl} showcases the creation of a PDDL plan derived from a functional unit.

%% file: 5-experiments.tex
\section{Experiments and Results}
The core aim of this experiment is to evaluate how our approach, which consolidates task plans, influences the accuracy and efficiency of planning. Using cooking as an illustrative example, we compared our model against other methodologies. We curated a dataset of 400 recipes, 5\% sourced from FOON and 95\% from Recipe1M+ \cite{marin2019learning}, which houses over a million recipes spanning various dish categories. Notably, the dataset included 100 recipes for each of the Salad, Drink, Muffin, and Omelette categories.

% Our findings are then compared to the results presented in \cite{Sakib2022ApproximateTT}.

\subsection{Evaluation Procedure}

\begin{figure}[!ht]
	\centering
        \includegraphics[width=\columnwidth]{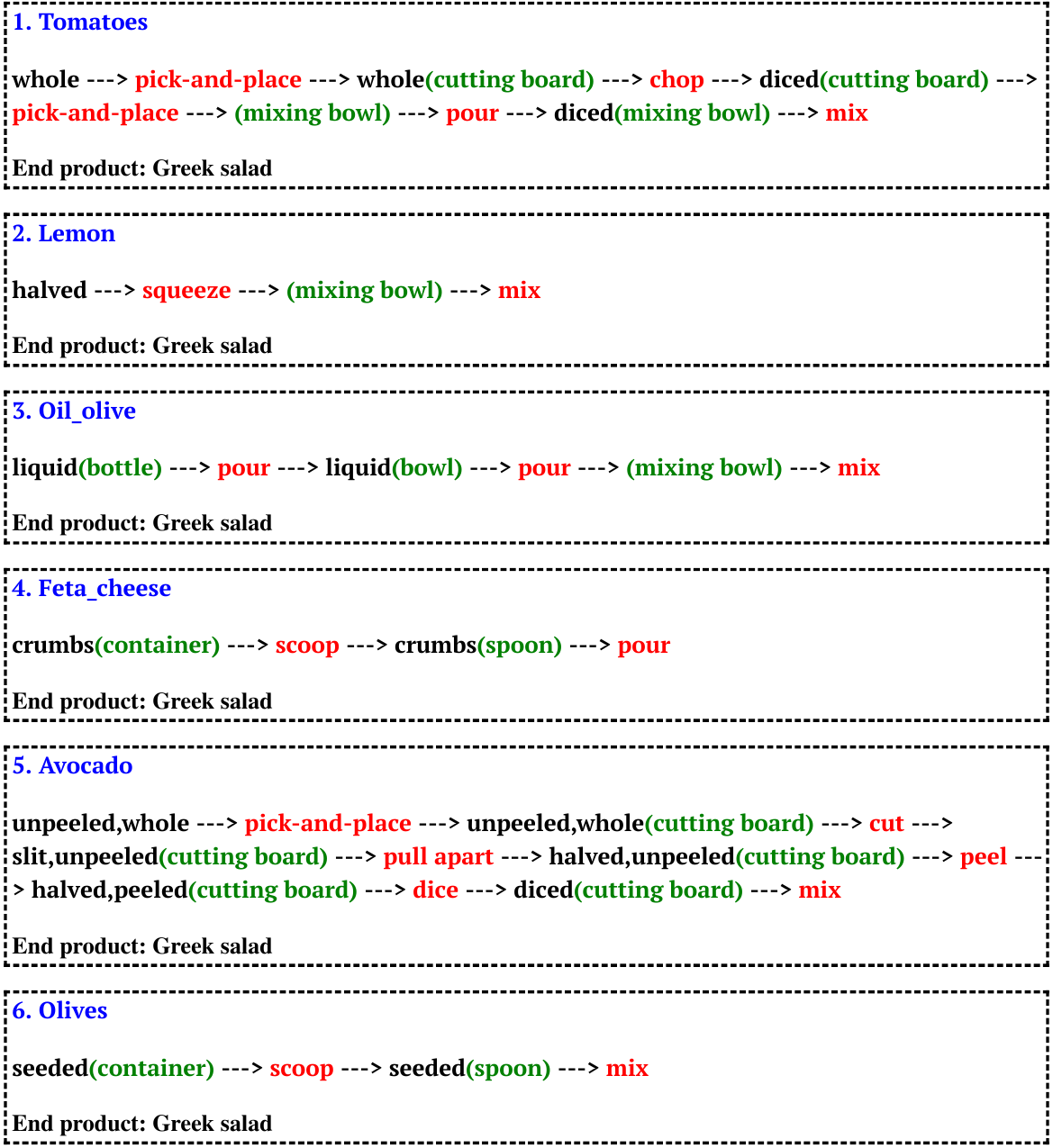}
	\caption{
        Progress lines for a Greek Salad recipe. The snapshot is taken from our task tree visualization tool.
	}
	\label{fig:progress_line}
\end{figure}

The verification of a cooking task plan by automated means presents a unique challenge due to the variability inherent in culinary preparation methods. Two different task plans representing the same dish may differ radically in their cooking techniques, but both could be considered valid. Consequently, this necessitates manual validation. However, the original format of a task tree can be difficult for humans to decipher. To mitigate this, we transform the task trees into progress lines, as demonstrated by Sakib et al. \cite{Sakib2022ApproximateTT} This helps illustrate the transformation and manipulation of ingredients throughout the cooking process. This streamlined visualization aids humans in spotting any inconsistencies in the task plan. A recipe is deemed accurate if the progress lines for all ingredients align correctly with the recipe steps.
An example of these progress lines as applied to a Greek Salad recipe is demonstrated in Figure \ref{fig:progress_line}.

\subsection{Task Planning Accuracy}
\begin{figure}[!ht]
	\centering
        \includegraphics[width=\columnwidth]{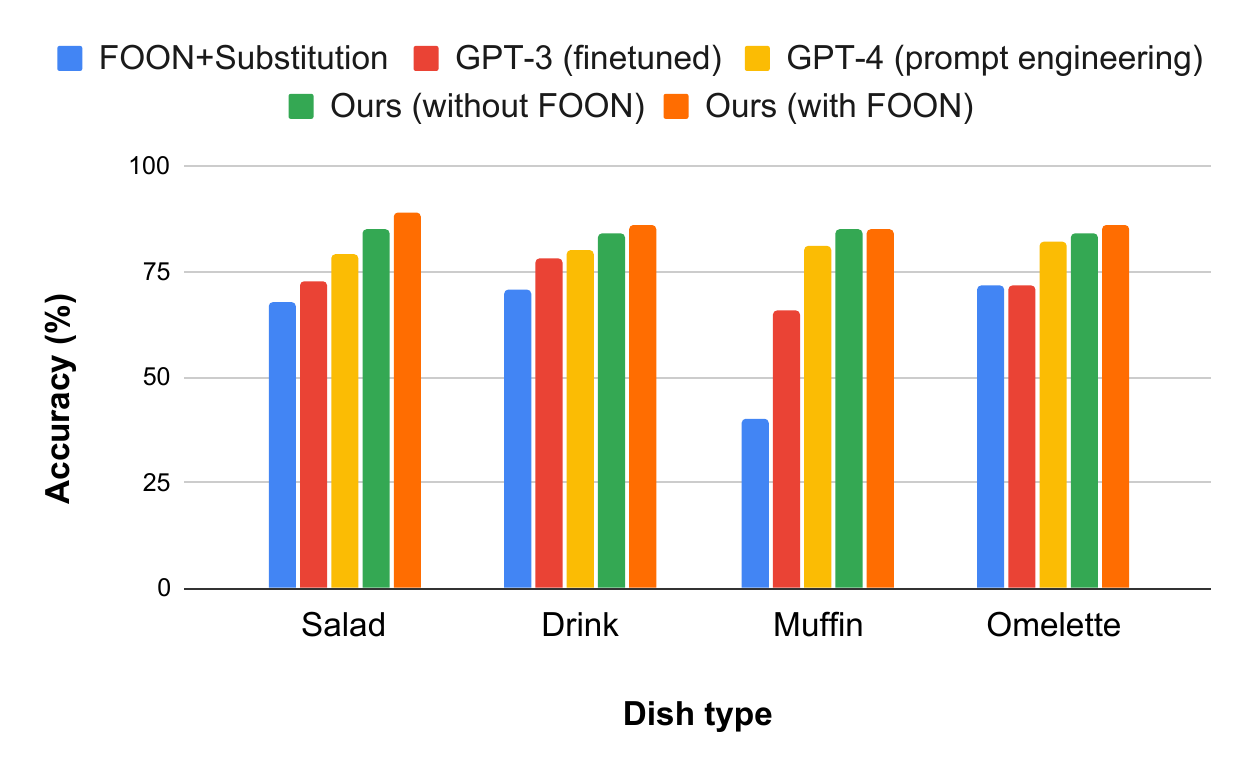}
	\caption{
        We employ cooking tasks as an example and evaluate various approaches' accuracy in generating plans for Salad, Drink, and Muffin dishes.
	}
	\label{fig:comparison}
       
\end{figure}

To evaluate the effectiveness of our devised task planning method, we conducted a comparative analysis with several established approaches. The results procured through progress line checking are displayed in Figure \ref{fig:comparison}.

\begin{itemize}
    % \item \textcolor{red}{LLM+P \cite{LLMP}: This technique is based on the premise that the knowledge base for a domain is pre-existing, and centers its focus on crafting the PDDL problem statement from natural language using LLMs. However, this claim cannot always hold true. For instance, in the field of cooking, an all-encompassing PDDL domain that can form a plan for every conceivable recipe is non-existent. Thus, we used FOON as the PDDL domain after suitable modifications, which acted as the foundational knowledge base for LLM+P implementation. Essentially, this transformation involved rendering each FOON functional unit into a PDDL planning operator. Since a functional unit's structure aligns closely with a PDDL action definition - both consisting of precondition, action, and effect - it validates using FOON as a feasible choice for conversion into a PDDL domain file. However, this approach could solely generate plans for the ones available in the domain file.}
    \item FOON+Substitution \cite{Sakib2022ApproximateTT}: This model does offer a degree of generalizability, but its effective operation is contingent upon the presence of a related recipe in FOON. This existing recipe is then manipulated to meet the necessary requirements.
    \item GPT-3 (finetuned): In our training process, we refined a GPT-3 model using recipes found in FOON. Initially, we commenced training with a sparse dataset of only 30 examples. This led to the model encountering difficulties in comprehending functional unit syntax, thus producing grammatical errors in the subsequent generation of functional units. For example, the model would incorrectly incorporateiple motion nodes within a single functional unit even though a functional unit, by definition, should contain only one motion node.

As we enriched the dataset with additional recipes, the model progressively rectified its syntactical inaccuracies. Nevertheless, it still demonstrated logical flaws such as improper state changes or missing actions. Figure \ref{fig:finetuning} demonstrates how the inclusion of new data to the training dataset incrementally led to improved model performance.

\begin{figure}[!ht]
	\centering
        \includegraphics[width=\columnwidth]{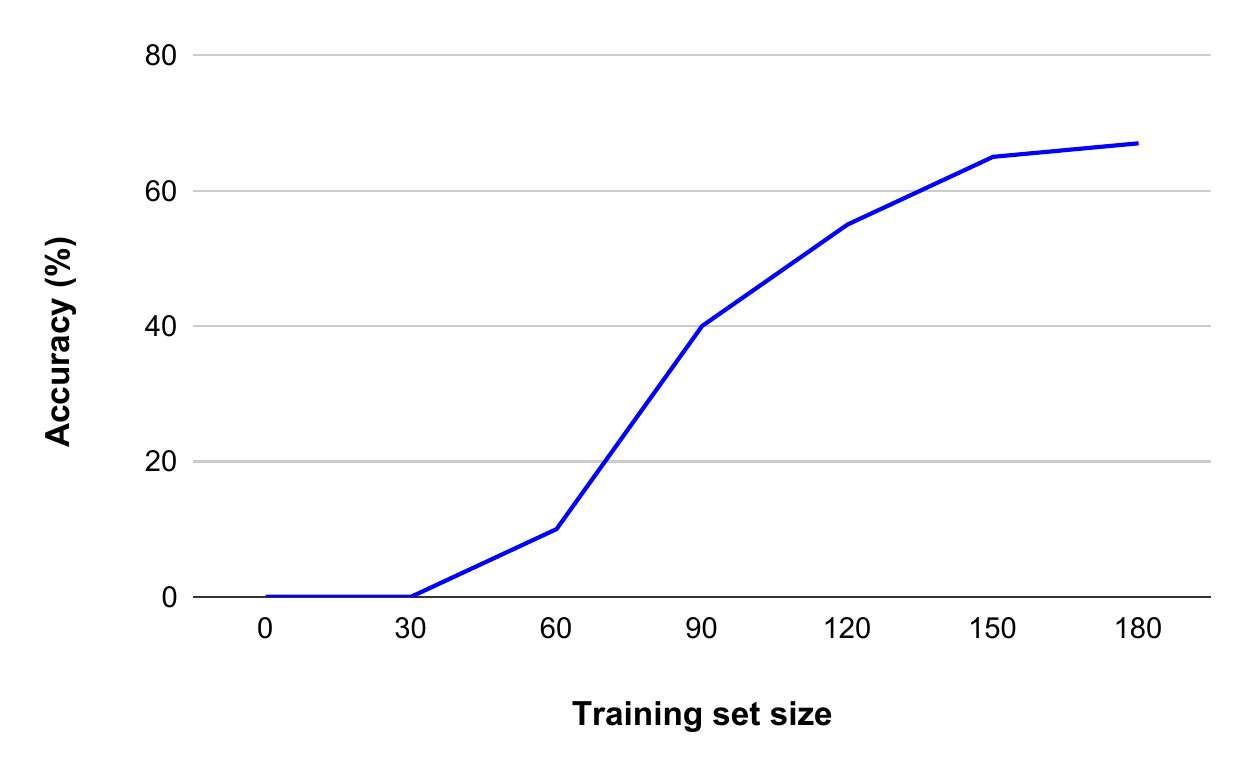}
	\caption{
        Impact of training dataset size on model accuracy.
	}
	\label{fig:finetuning}
\end{figure}
    
    \item GPT-4 (prompt engineering): This method utilizes prompt engineering with GPT-4, excluding the consolidation of task trees.

\end{itemize}

The FOON-search approach heavily relies on finding a similar recipe within FOON as a reference for modifying the task plan. Consequently, a substantial number of adjustments were necessary, resulting in inaccuracies within the task plan. This is particularly noticeable in the higher accuracy achieved for Salad in comparison to Muffin, attributed to the limited number of Muffin recipes available in FOON.

% \textcolor{red}{LLM+P showcased lower accuracy due to its reliance on existing recipes within FOON, which constituted only 5\% of the dataset used in our experiment. Consequently, this method was successful solely for recipes available within FOON.}

The findings demonstrate that the LLM-based approach does not exhibit a dependency on the recipe type. Our GPT-4-driven method surpassed the fine-tuned GPT-3 approach in performance. Further enhancements were realized by integrating FOON into the pipeline.

\subsection{Cost Optimization}

\begin{figure}[!ht]
	\centering
        \includegraphics[width=\columnwidth]{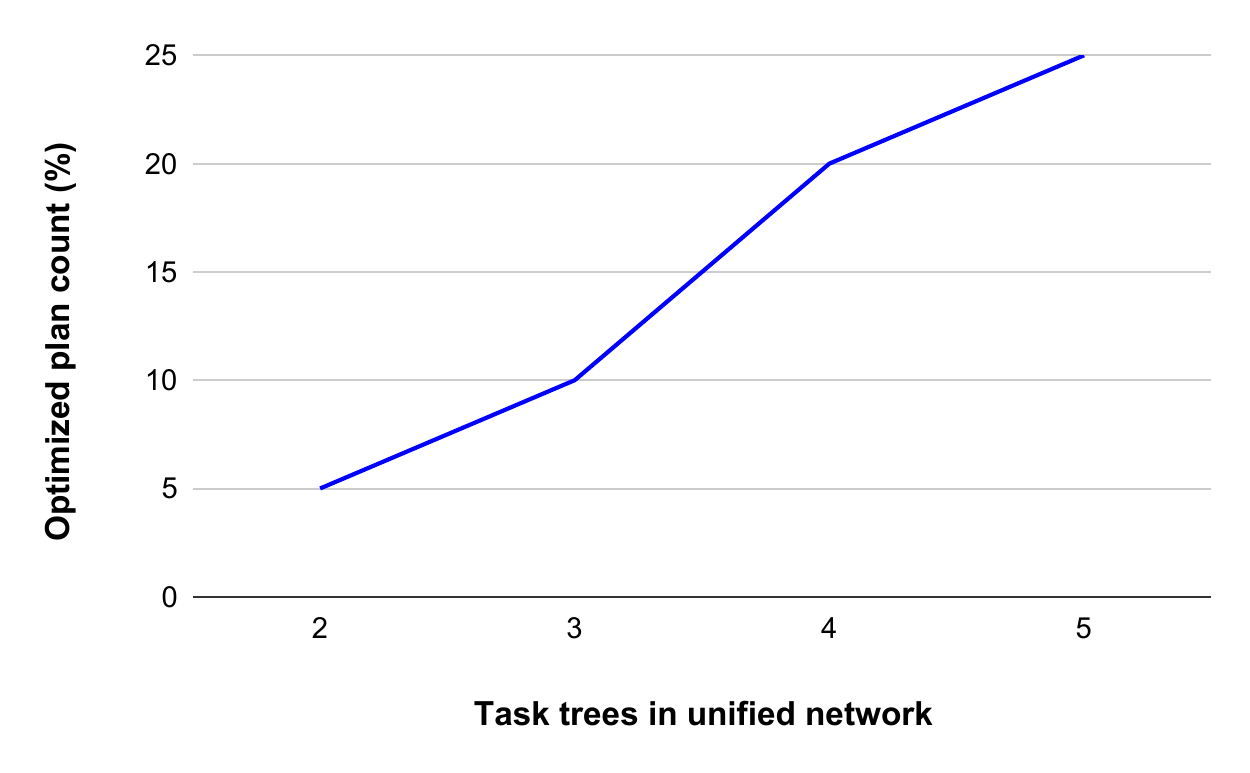}
	\caption{
        X-Axis: This represents the quantity of task trees produced for a particular task. They were then unified to establish a comprehensive network. Y-Axis: This shows the count of recipes that have been optimized when compared to the initial action plan generated by GPT-4. 
	}
	\label{fig:cost}
\end{figure}

The objective of this experiment is to evaluate the extent to which our approach can optimize a task plan. In the context of cooking, if a recipe cannot be optimized, it implies that there are no superior alternatives compared to the initial output generated by LLMs. In Figure \ref{fig:cost}, we present the number of optimized recipes by generating different numbers of task trees and then unifying them in a network. When there are only 2 task trees merged in the network, it yields a better solution compared to the first generated tree in 5\% of the cases. Similarly, by gradually increasing the number of task trees in the network up to 5, we obtain a more cost-effective solution in 25\% of the cases. 
The results suggest that increasing the quantity of task trees merged into the network can enhance the optimization of task plan. For tasks with more options and flexibility, the possibility of finding an optimized solution is higher. 

%More optimization occurs when we choose task tree 6 from the Mini-FOON, as it combines subtasks from five different task trees, resulting in a lower cost. Ultimately, task tree 7, the final output from our pipeline, maximizes the advantages of FOON and minimizes the execution cost compared to task tree 1 in 40\% cases. 

% This experiment provides evidence that our method generates plans that are more likely to be successfully executed by the robot.

% \begin{wrapfigure}{r}{0.6\textwidth}
%   \begin{center}
%     \includegraphics[width=0.58\textwidth]{figures/cost_optimization.pdf}
%   \end{center}
%   \caption{The graph depicts the number of recipes that were optimized by generating varying numbers of task trees in comparison to Task Tree 1} 
%   % The performance improvement is evaluated as we gradually increase the size of the finetuning dataset.}
%   \label{fig:cost}
  
% \end{wrapfigure}

% \begin{figure}[ht]
% 	\centering
% 	\includegraphics[width=0.8\columnwidth]{figures/cost_optimization.pdf}
% 	\caption{The graph depicts the number of recipes that were optimized by generating varying numbers of task trees in comparison to Task Tree 1. Increasing the number of task trees improves the probability of obtaining a more efficient plan. 
% }
% 	\label{fig:cost}
%         \vspace{-0.2in}
% \end{figure}

%% file: 6-discussion.tex
\section{Discussion} 

\subsection{Generalization Capability}
\begin{figure}[ht]
	\centering
	\includegraphics[width=\columnwidth]{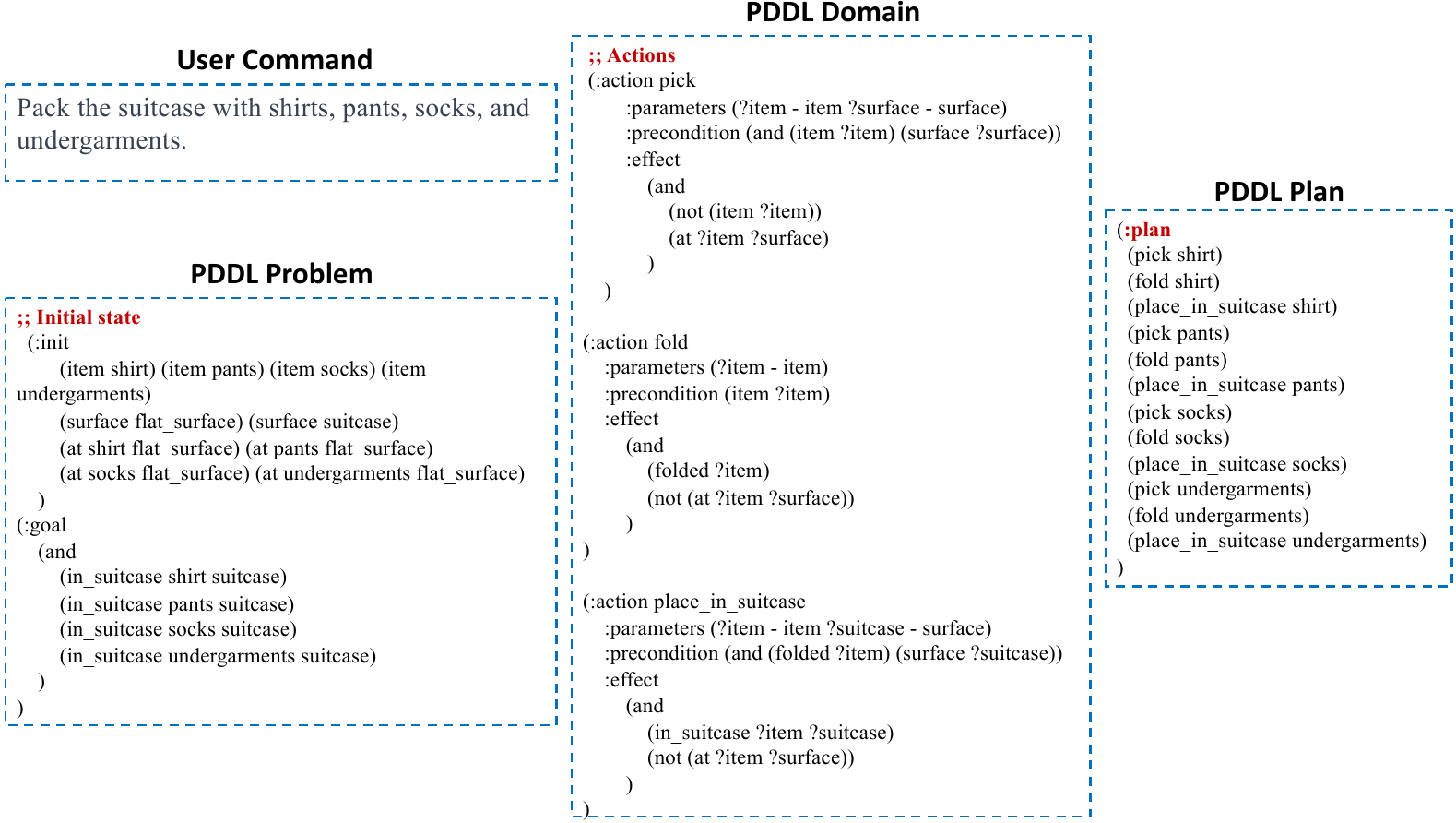}
	\caption{Generation of PDDL domain with the example of problem and the task plan using our pipeline for packing up a suitcase.
}
	\label{fig:suitcase}
\end{figure}

Our approach can be seamlessly adapted to various domains with little or no component adjustments. For example, the depicted scenario in Figure \ref{fig:suitcase} involves instructing a robot to pack a suitcase. %Since no existing knowledge base like FOON is available for this domain, 
Our approach constructs a unified network solely from LLM-generated task trees and identify an optimal plan from the available alternatives. 

\subsection{Plan Visualization and Correction} 
\label{human}
Our method offers an additional unique advantage by providing a visual representation of the progress line of each key object throughout the task plan, following the generation of an optimal task tree. These progress lines enable users to review and make adjustments or corrections whenever an error is identified. The corrected tree is then converted into a PDDL plan. This feature allows regular users to modify the plan while it is still easily understandable, as PDDL code is not typically comprehensible to regular users. Human intervention in the task tree format using the visualization tool can significantly enhance the likelihood of successful robotic execution. Moreover, this visual depiction facilitates understanding of the robot's decision-making process, offering insights into its reasoning.

An illustration of the plan visualization can be seen in Figure \ref{fig:progress_line}. For instance, if a user observes that the olive oil (depicted as the third ingredient in the figure) is initially placed in a cup according to the actual scenario, but the plan illustrates it to be in a bottle, this discrepancy can be corrected in the visualization tool. For that, user simply needs to edit the text shown as the progress line. The adjustment will then be automatically reflected back to the task tree.

\subsection{Future Directions}
Robots can encounter failures while trying to complete tasks due to factors such as incorrect task or motion plans, system malfunctions, and more. For example, a robot attempting to grasp an object may experience it slipping and falling. It is crucial to develop a way for robots to recover from such failures autonomously, without the need for human intervention.
In our future investigations, we aim to enhance the system's adaptability to autonomously address plan failures. To facilitate this, we recognize the importance of environmental feedback. If the robot utilizes its visual system to capture real-time environmental data and creates a corresponding PDDL problem, it can rectify plan errors without human intervention. While GPT-4 supports image input capability, it is currently inaccessible to the public. Our upcoming experiments will focus on whether the robot, with its vision system and GPT-4, can accurately identify and update the environmental state.

In the future, we plan to adjust our approach for other types of tasks that inherently lack alternative solutions. For instance, assembling a specific model of an IKEA chair typically has a limited number of solutions. For such tasks, the generalization capability of LLMs might not be necessary. In our future studies, we intend to explore methodologies to determine the potential benefits of generating multiple task trees for a given task.

Integration of the Tree-of-Thoughts (ToT) \cite{yao2023tree} approach into our current pipeline could present an exciting new development. ToT allows LLMs to assess various reasoning routes at each intermediary stage and determine the path most closely aligned with the end goal. The method has demonstrated significant advancements in domains like crossword puzzles and creative writing. However, to apply this to an alternative domain, it requires a way for LLMs to independently evaluate the different choices at each juncture. Our future investigations will focus on whether this concept can be appropriately adapted for generating task trees for long-horizon tasks, such as cooking, furniture assembly, and manufacturing.

%% file: 7-conclusion.tex
\section{Conclusion}

% In this study, our objective was to propose a novel pipeline for task tree generation, leveraging the advantages offered by LLMs. We utilized ChatGPT to respond to user queries, and then fine-tuned a GPT-3 model to convert the response into a task tree representation. To enhance the accuracy and execution cost of the task tree, we integrated the output of the fine-tuned model with FOON, exploring multiple possibilities to achieve the desired objectives. Through our experiments, we demonstrated its superior performance, highlighting its remarkable generalization capabilities.
% In future, we intend to focus on addressing the challenges of task tree correction and re-planning in cases of planning or execution failures. It is worth noting that our pipeline exhibits a high degree of flexibility, allowing for the seamless substitution of GPT and FOON with more advanced Language Models or knowledge networks. We aim to incorporate image inputs into our system by utilizing the newly released GPT-4, which can handle both textual questions and accompanying images. This would allow users to upload images of dishes and inquire about their preparation methods.

This study introduces a pipeline aimed at generating accurate and optimal robotic task plans across various problems, focusing on enhancing the reliability of plans generated by LLMs. In contrast to conventional methods producing a single plan, our innovative approach, utilizing GPT-4, yields multiple high-level task plans structured as task trees based on natural language instructions.
These task trees are amalgamated into a cohesive network by eliminating unreliable and costly elements, significantly enhancing planning accuracy and task execution efficiency. Subsequently, we employ GPT-4 to translate the task tree into an executable PDDL plan, optimized for robotic system implementation.
The efficacy of this method is verified through comprehensive experiments, particularly in the context of planning tasks for a cooking robot. Moving forward, our focus will be on addressing the challenges of re-planning in cases of planning or execution failures.